\documentclass[sigconf,screen]{acmart}
\AtBeginDocument{%
  \providecommand\BibTeX{{%
    \normalfont B\kern-0.5em{\scshape i\kern-0.25em b}\kern-0.8em\TeX}}}

\setcopyright{acmcopyright}
\copyrightyear{2018}
\acmYear{2018}
\acmDOI{XXXXXXX.XXXXXXX}

\acmConference[Conference acronym 'XX]{Make sure to enter the correct
  conference title from your rights confirmation emai}{June 03--05,
  2018}{Woodstock, NY}
\acmPrice{15.00}
\acmISBN{978-1-4503-XXXX-X/18/06}

\begin{document}

\title{Uplift Modeling based on Graph Neural Network Combined with Causal Knowledge}

% Optional math commands from https://github.com/goodfeli/dlbook_notation.
% \input{math_commands.tex}

% \usepackage{hyperref}
% \usepackage{url}
% \usepackage{graphicx}

\author{Haowen Wang}
\affiliation{%
  \institution{Alipay, AntGroup}
  \country{China}
}
\email{wanghaowen@antgroup.com}

\author{Xinyan Ye}
\affiliation{%
  \institution{Imperial College London}
  \country{UK}
}
\email{xy2119@ic.ac.uk}

\author{Yikang Wang}
\affiliation{%
  \institution{University College London}
  \country{UK}
}
\email{yikang.wang.21@ucl.ac.uk}

\author{Yangze Zhou}
\affiliation{%
  \institution{Zhejiang University}
  \country{China}
}
\email{yangze.zhou@zju.edu.cn}

\author{Zhiyi Zhang}
\affiliation{%
  \institution{Peking University}
  \country{}
}
\email{emma0302@pku.edu.cn}

\author{Longhan Zhang}
\affiliation{%
  \institution{Zhejiang Lab}
  \country{China}
}
\email{longhanz@zhejianglab.com}

\email{emma0302@pku.edu.cn}

\author{Jing Jiang}
\authornote{Corresponding Author.}
\affiliation{%
  \institution{Zhejiang Lab}
  \country{China}
}
\email{jiangj@zhejianglab.com}

\renewcommand{\shortauthors}{Wang and Ye, et al.}

\begin{abstract}
Uplift modeling is a fundamental component of marketing effect modeling, which is commonly employed to evaluate the effects of treatments on outcomes. Through uplift modeling, we can identify the treatment with the greatest benefit. On the other side, we can identify clients who are likely to make favorable decisions in response to a certain treatment. In the past, uplift modeling approaches relied heavily on the difference-in-difference (DID) architecture, paired with a machine learning model as the estimation learner, while neglecting the link and confidential information between features. We proposed a framework based on graph neural networks that combine causal knowledge with an estimate of uplift value. Firstly, we presented a causal representation technique based on CATE (conditional average treatment effect) estimation and adjacency matrix structure learning. Secondly, we suggested a more scalable uplift modeling framework based on graph convolution networks for combining causal knowledge. Our findings demonstrate that this method works effectively for predicting uplift values, with small errors in typical simulated data,  and its effectiveness has been verified in actual industry marketing data.
\end{abstract}

\begin{CCSXML}
<ccs2012>
<concept>
<concept_id>10002951.10003227</concept_id>
<concept_desc>Information systems~Information systems applications</concept_desc>
<concept_significance>500</concept_significance>
</concept>
<concept>
<concept_id>10010147.10010178</concept_id>
<concept_desc>Computing methodologies~Artificial intelligence</concept_desc>
<concept_significance>500</concept_significance>
</concept>
<concept>
<concept_id>10010147.10010257</concept_id>
<concept_desc>Computing methodologies~Machine learning</concept_desc>
<concept_significance>500</concept_significance>
</concept>
</ccs2012>
\end{CCSXML}

\ccsdesc[500]{Information systems~Information systems applications}
\ccsdesc[500]{Computing methodologies~Artificial intelligence}
\ccsdesc[500]{Computing methodologies~Machine learning}

\keywords{Uplift Modeling, Graph Neural Network, Causal Inference}

\maketitle

\section{Introduction}
Uplift modeling~\citep{uplift_modeling} has traditionally relied on randomized experiments, such as randomized controlled trials (RCTs)~\citep{RCT}, in which customers are randomly allocated to either receive or not receive the intervention. In such instances, obtaining an accurate and interpretable estimate from observational data becomes critical. However, carrying out such an experiment in a business context frequently results in several challenges, including high costs in terms of time and money, uneven intervention distribution, and selection bias in the specific population.

Response modeling or outcome prediction uses supervised learning models to model the relation between features and target variables to predict response variation. Although response modeling is typically preferable to random targets, distinguishing between treatment-induced be- behavioral changes is often challenging. The population that should be targeted is the one most likely to respond positively to the intervention. As a result, a thorough knowledge of the behavioral changes that occur after the intervention is essential. Uplift modeling simulates the causal effect between the intervention and the outcomes based on response modeling. Causal inference frameworks and machine learning models are corporated to provide accurate forecasts and optimized performance on intuitive metrics.

The counterfactual nature of intervention data is central to causal inference in Rubin's Potential Outcome Framework (POF)~\citep{rubin2005causal}. This characteristic pertains to a person's inability to both receive and refuse intervention. This means that the effects of many therapies cannot be seen in the same person. Two frameworks that have been extensively examined for causal impact estimations based on this counterfactual characteristic are the meta-learner framework~\citep{meta:Ton} and the customised machine learning model-based framework~\citep{ML:Chen}. The ultimate goal is to increase the accuracy of causal impact estimation through the use of feature engineering and validation approaches such as PS matching~\citep{PSM:Caliendo}, weighting~\citep{weighting:Li}, feature representation~\citep{fe:Muandet}, and so on.

In the past, researchers in uplift modeling were largely concerned with how to employ unbiased data and models in the estimation framework. We increased the amount of data information by defining causal knowledge and implementing structured representation, then used a graph convolution neural network~\citep{GCN:Zhang}  to efficiently and directionally integrate feature neighborhood information, achieving excellent performance in uplift modeling and prediction tasks. The following is a description of our paper's contribution to methodological and empirical evaluation perspectives:

$\bullet$
First, we propose to use conditional average treatment effect (CATE) as the attribute representing the causal information of the feature and as part of uplift modeling and propose a causal network model framework to effectively calculate it based on knowledge distilling and double machine learning.

$\bullet$ 
Second, we propose to learn the causal diagram structure of the data before uplift modeling and reconstruct the data according to the learned adjacency matrix.

$\bullet$ 
Third, we propose an uplift modeling estimator based on graph convolution neural networks, which can integrate and characterize neighborhood feature attributes according to the cause and effect diagram structure and improve the performance of downstream tasks.

\section{Related Work}

The estimation of the uplift value in uplift modeling is often based on the Potential Outcome Framework(POF)~\citep{rubin2005causal}. The individual treatment effect(ITE) can be expressed as:
\begin{equation}
I T E:\tau(i) = Y_{i}(1) - Y_{i}(0)
\end{equation}
Shere $Y_{i}(1)$ and $Y_{i}(0)$ represents the result of the outcome variable under the treatment condition and control condition, respectively, for individual $i$, $\tau(i)$ is the ITE value.

Considering that the individual effect of treatment will vary from individual and the high cost of marketing experiments in the industry, the conditional average treatment effect(CATE) is proposed as the effect of treatment on subgroups evaluated by the conditional average treatment effect (CATE), which is calculated by:
\begin{equation}
C A T E: \tau_{i}=E\left[Y_{i}(1) \mid X_{i}\right]-E\left[Y_{i}(0) \mid X_{i}\right]
\end{equation}
where $X_i$ is the feature vector for individual $i$.

For the estimation of CATE and ITE, the most direct method is to make an unbiased adjustment to the regression model. Series of meta learners represented by s-learner are designed based on the concept, that is, train one or more models with y as the output training target, input T and X, and get the change of Y by changing the value of T to estimate ITE and CATE.

\begin{equation}
\tau(x) = E[Y_i(1) - Y_i(0)|X] = E[\tau_i|X]
\end{equation}

Another series of methods for uplift modeling prediction is the probability score matching (PSM)~\citep{PSM:Caliendo} method based on randomized controlled trials (RCT)~\citep{RCT}. By calculating the probability score $P(t \mid x)$, each sample is given a different treatment object according to its similarity, so for sample $i$, we find sample $j$:

\begin{equation}
    \operatorname{argmin}_{j} \operatorname{dist}(i, j)=\left|P\left(t \mid x_{i}\right)-P\left(t \mid x_{j}\right)\right|
\end{equation}

Then CATE could be calculated by:

\begin{equation}
    \hat{\tau}=\frac{1}{n}\left[\sum_{i: t_{i}=1}\left(y_{i}-y_{j}\right)+\sum_{i: t_{i}=0}\left(y_{j}-y_{i}\right)\right]
\end{equation}

In addition, the industry's research on lift modeling also includes methods based on the Covariate Balancing Method and Modeling Unobserved Confounder. Typical methods of the first category include Inverse Probability of Treatment Weighting (IPTW)~\citep{IPTW:Chesnaye}, Entropy Balancing (EB)~\citep{EB:Hainmueller}, and Approximate Residual Balancing (ARB)~\citep{ARB:Athey}, in which core is how to re-assign weights to samples. The core of the second type of method is to model the confounder. One way is to model the instrumental variable, which is represented by the two-stage least square (2SLS) method~\citep{2SLS:Bollen}. The first stage is to fit the impact of the change of I on T, and the second stage is to fit the impact of the change of T on y caused by the change of I. The other way is to use deep learning to represent the confounder, such as SITE~\citep{SITE:Yao}, Dragonnet~\citep{DragonNet:Tso}, and CEVAE~\citep{CEVAE:Louizos}.

The past research mainly focused on adjusting and optimizing the uplift value estimation model in a structured or unstructured way. Estimation methods based on the foundation model have shown us the importance of embedding causal structure knowledge into the estimation process. This paper will try to conduct data mining on features. On the one hand, it expands the amount of information by defining and applying causal information; on the other hand, it reconstructs structured origin data through causal diagram structural information and uses GCN to learn neighborhood information from unstructured reconstructed data to improve the performance of uplift modeling using the framework of meta learner~\citep{meta:Ton}.

The structure of the paper is as follows. Section II reviews the critical concepts of uplift modeling and frameworks of the learning approach. In Section III, we introduce the methodology of our causal knowledge framework. Section IV evaluates these methods with both synthetic and real-world data. Finally, Section V summarizes the findings and recommends future research for uplift modeling applications.

\section{Methodology}

This section will introduce the calculation method and architecture of graph neural networks embedded with causal knowledge.
We propose an interpretable causal graph network representation learning framework with features as nodes. It can expand the representation of features by node embedding, mapping the originally scalar features into a high-dimension space, and then integrating the causal information and structural information into the graph features through graph convolution to achieve a more accurate estimation of uplift value.

\subsection{Causal Knowledge Representation}
We propose a framework for computing causal knowledge representation. We transfer knowledge through the concept of the soft target in knowledge distillation as the estimation target of the causal estimator. We estimate each feature's causal average treatment effect(CATE) and take it as the weight of the feature based on the causal effect. This work has been proven to obtain more information.

\begin{figure}[h]
\centering
\begin{minipage}[t]{0.48\textwidth}
\centering
\includegraphics[width=6cm]{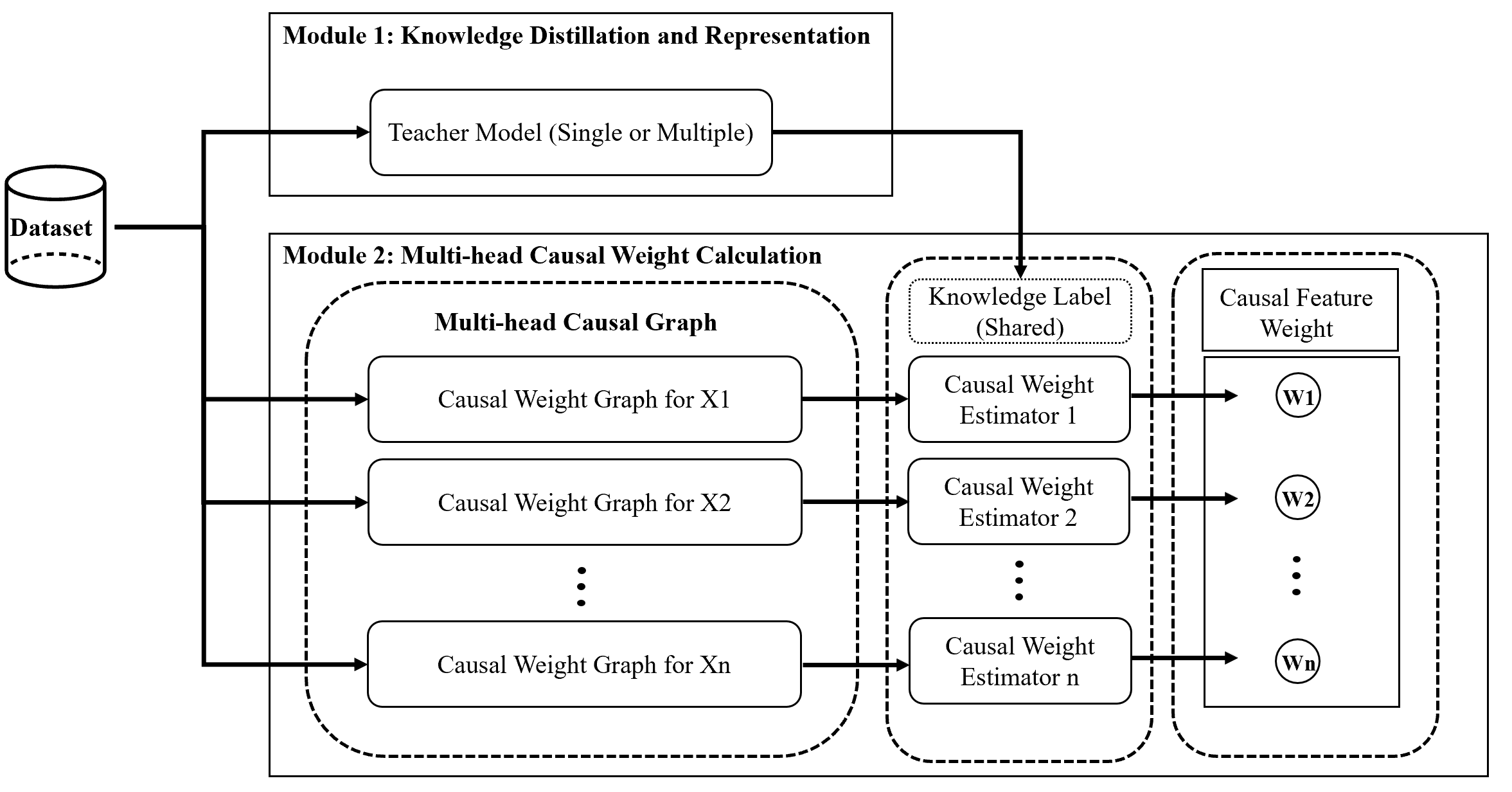}
\caption{Causal Weighting Calculation Framework}
\label{causal weighting cal}
\end{minipage}
\begin{minipage}[t]{0.48\textwidth}
\centering
\includegraphics[width=5.5cm]{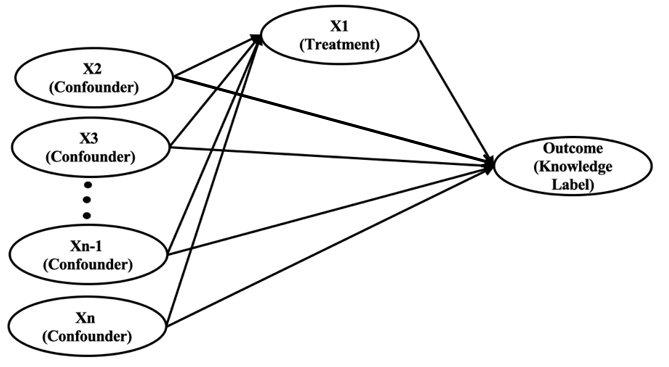}
\caption{Causal Graph for Feature X1}
\label{causal graph}
\end{minipage}
\end{figure}

Figure~\ref{causal weighting cal} shows the architecture of the causal average treatment effect(CATE) calculation. Firstly, in the module of knowledge distillation and representation, We will build a knowledge distillation task for label Y(0/1), using the teacher model(XGBoost~\citep{xgboost:Chen}, etc. as base regressor) to get the probability $\hat{Y}$ as the soft label to replace Y as the target label. Secondly, in the multi-head causal weight calculation module, we establish a causal graph for each feature as shown in Figure~\ref{causal graph}, use the soft label $\hat{Y}$ got in Module 1 as a knowledge label, and estimate CATE in the framework of double machine learning (DML)~\citep{DML:Chernozhukov}. Since the CATE estimation of each feature is independent, we designed a multi-head mechanism to make the calculation more efficient. Double machine learning is a classic estimator to estimate (heterogeneous) treatment effects when treatment is classified and all potential confounders/controls. DML makes the following structural equation assumptions for the data generation process:

\begin{equation}
Y=\theta(X) \cdot T+g(X, W)+\epsilon \quad \mathbb{E}[\epsilon \mid X, W]=0
\end{equation}

\begin{equation}
T=f(X, W)+\eta \quad \mathbb{E}[\eta \mid X, W]=0
\end{equation}

\begin{equation}
\mathbb{E}[\eta \cdot \epsilon \mid X, W]=0
\end{equation}

After modeling Y and T, respectively, the estimated CATE value $\theta(X)$ satisfies the equation:

\begin{equation}
    \tilde{Y}=\theta(X) \cdot \tilde{T}+\epsilon
\end{equation}
Here $\tilde{Y}$ is the residual of Y, $\tilde{T}$ is the residual of T.

Considering  $\mathbb{E}[\epsilon \cdot \eta \mid X]=0$, the problem of estimating  $\theta(X)$ can be transformed into the following regression problem.

\begin{equation}
    \hat{\theta}=\arg \min _{\theta \in \Theta} \mathbb{E}_{n}\left[(\tilde{Y}-\theta(X) \cdot \tilde{T})^{2}\right]
\end{equation}

\subsection{Causal Graph Structure Learning}
The graph network structure contains the dataset's prior information. The connection relationship indicates the direction and distance of information transmission and determines the direction and degree of information sharing and transmission of nodes in the subsequent graph network characterization operation.

Here, we use the classical Bayesian network structure as the structure of the causal feature representation graph. Scoring search is a standard method to solve the problem of Bayesian network structure to evaluate the degree of fit between the Bayesian network and training data and then find the optimal Bayesian network based on the scoring function. The goal is now to solve the following task:
\begin{equation}
    \underset{G \in G}{\operatorname{argmax}} \operatorname{score}(G, \mathcal{D}) .
\end{equation}
The scoring function introduces the inductive preference of what kind of Bayesian network you want to obtain. Here we use the Bayesian Information Criterion(BIC)~\citep{BIC:Neath} as the score function, which approximates the Bayes Dirichlet equivalent uniform(BDeu), sharing the critical property of decomposability.
\begin{equation}
    \operatorname{score}(G, \mathcal{D})=\sum_{X_{i}} \operatorname{score}\left(X_{i}, \Pi_{i}, \mathcal{D}\right)
\end{equation}
\begin{equation}
    \operatorname{Score}_{b i c}(g: D)=l((\hat{\theta}, g): D)-\frac{\log M}{2} \operatorname{Dim}[g]
\end{equation}

D is the given data, M is the number of training samples, g is the given structure, Dim[g] is the number of independent parameters of model g, $\hat{\theta}$ is the maximum likelihood estimate of the parameter given the structure g and the data D.

After determining the scoring function, here we use the hill-climbing~\citep{hill:Selman} algorithm as the optimization algorithm for the structural learning problem. The Hill-climbing algorithm is a classical algorithm for local search based on a greedy algorithm, starting with a candidate solution and continuing to search in its neighborhood until there is no better solution. The steps of the local search algorithm are described as follows: Firstly, initialize a feasible solution X. Secondly, select a moved solution s (x) in the neighborhood of the current solution so that f (s (x)) < f (x), s (x) \ in S (x). If there is no such solution, X is the optimal solution, and the algorithm stops. Thirdly, make x = s (x) and repeat the second step.

\subsection{GNN based uplift modeling}

After Causal Knowledge Representation and Causal Graph Structure Learning, we obtained more information about the dataset and a specific relationship between features. Considering the excellent representation ability of graph neural networks, we propose a graph neural network representation framework based on causal graph representation, which can integrate this information more efficiently.

GCN~\citep{GCN:Zhang} is a multi-layer neural network that can operate directly on the graph and induce nodes to obtain information on neighborhood vectors based on the neighborhood attributes of nodes. Consider a graph G = (V, e), where V (| V | = n) and E are the sets of nodes and edges, respectively. 
It is assumed that each node is connected to itself, that is, (v, v) $\in E$ for any v. Let x $\in \mathbb{R}^{n \times m}$ be a matrix containing all N nodes and their vector features, where m is the dimension of the vector, and each row $x_{v} \in \mathbb{R}^{m}$ are the vectors of V. We introduce the adjacency matrix A and its degree matrix D of G, where $d_{i i}=\sum_{j} A_{i j}$. Due to the characteristics of the self-circulation hypothesis, the diagonal element of a is set to 1. In general, GCN can only capture information about its neighbors through one layer of convolution. We can integrate information about a wider range of neighbors by stacking multiple GCN layers:

\begin{equation}
    H^{(l+1)}=\sigma\left(\hat{D}^{-\frac{1}{2}} \hat{A} \hat{D}^{-\frac{1}{2}} H^{(l)} W^{(l)}\right)
\end{equation}

Here $H^{(l+1)}$ and $H^{(l)}$ are the output and input matrices. $\hat{A}=A + I$, where A is the adjacency matrix, and I is the identity matrix. $\hat{D}$ is the degree matrix of $\hat{A}$, $\hat{D}^{-\frac{1}{2}} \hat{A} \hat{D}^{-\frac{1}{2}}$   is the normalized symmetric adjacency matrix, and  $W^{(l)} \in \mathbb{R}^{m \times k}$  is a weight matrix. $\sigma$  is an activation function, e.g., a  $\operatorname{LeakyReLU}$.

\begin{figure}[h]
  \centering
  \includegraphics[width=\linewidth]{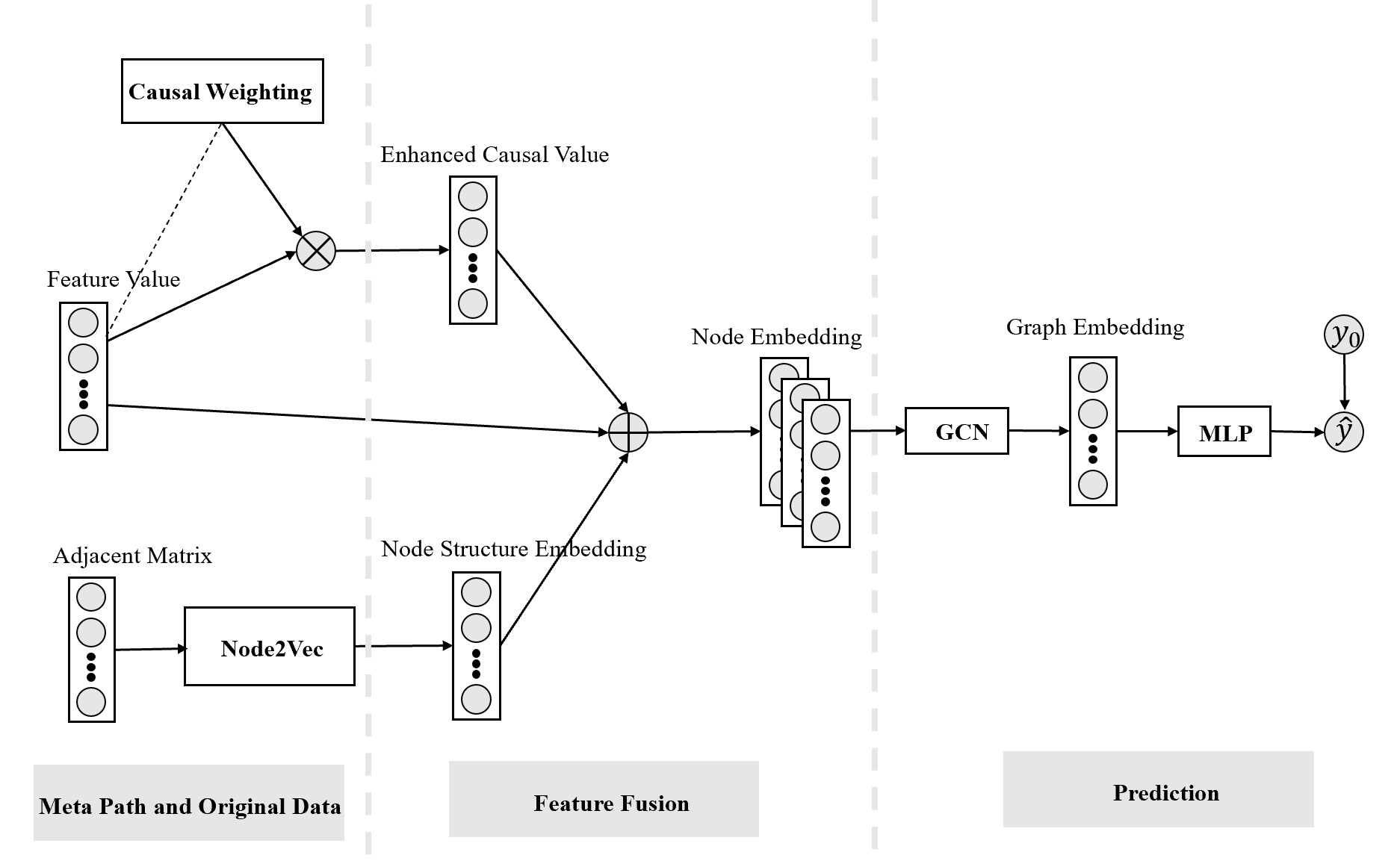}
  \caption{GNN-based uplift modeling architecture.}
  \label{uplift_GNN}
\end{figure}

Here we use GCN to extract and integrate features based on the causal neighborhood structure we learned in the previous step. We take advantage of the feature that GCN can efficiently fuse features according to the neighborhood structure to get graph embedding of each sample and then perform prediction tasks based on it. Figure~\ref{uplift_GNN} shows that we have expanded the information on each feature. In addition to the value of each feature itself, we have also expanded the information of structural features and causal weights.

For the estimation of uplift value, we refer to the design method of S-learner in meta learner and use our GNN-based model as the base learner. 

\begin{equation}
     \mu_{0}(x)=\mathbb{E}[Y(T=0) \mid X=x]
\end{equation}
\begin{equation}
     \mu_{1}(x)=\mathbb{E}[Y(T=1) \mid X=x] 
\end{equation}

After $\mu_{0}$ and $\mu_{1}$ are calculated, respectively, the uplift value for each sample can be calculated:

\begin{equation}
     \tilde{D}_{i}^{1}:=Y_{i}^{1}-\hat{\mu}_{0}\left(X_{i}^{1}\right)
\end{equation}
\begin{equation}
     \tilde{D}_{i}^{0}:=\hat{\mu}_{1}\left(X_{i}^{0}\right)-Y_{i}^{0} 
\end{equation}

Here $\tilde{D}_{i}^{1}$ and $\tilde{D}_{i}^{0}$ are the uplift values for samples in the intervention group and control group.

\section{Experiments}
\subsection{Dataset}
\subsubsection{Synthetic dataset}

We used a method to simulate the generation of a dataset containing individual treatment effects, which is available in causalml. In \citet{ML:Chen} research,  it is used as a method to provide simulated data, which is available in Causalml. This synthetic method in the study provides the test groundings for estimating individual treatment effects and facilitating validation. The following is the generating mechanism:  for different choices of X-distribution $P_{d}$, there is dimension $d$, noise level $\sigma$, propensity function $e^\ast(\cdot)$, baseline primary effect $b^\ast(\cdot)$, and treatment effect function $\tau^\ast(\cdot)$. The distributions and relations are mathematically expressed in terms: 

\begin{equation}
    X_{i} \sim P_{d}
\end{equation}

\begin{equation}
    \varepsilon_{i} \mid X_{i} \sim N(0,1)
\end{equation}

\begin{equation}
    W_{i} \mid X_{i} \sim \text { Bernoulli }\left(e^{*}\left(X_{i}\right)\right)
\end{equation}

\begin{equation}
    Y_{i}=b^{*}\left(X_{i}\right)+\left(W_{i}-0.5\right) \tau^{*}\left(X_{i}\right)+\sigma \varepsilon_{i}
\end{equation}

The generation mechanism is featured by nuisance components  and an easy treatment effect function.  The initial distribution is constructed from $X_{i1} \sim Unif(0, 1)^d$ is constructed, followed by $e^\ast(X_{i} = trim_{0.1}\{sin(\pi X_{i1}X_{i2})\}$ and $\tau^\ast(X_{i}) = (X_{i1}+X_{i2})/2$ to compute propensity scores and treatment effects, respectively. The treatment $W$ is generated as a binary distribution. Eventually, interval trimming of the distribution is performed via $trim (x) = max\{\eta, min(x, 1 -\eta)\}$. This simulation method is adpoted as a scaled version of the Friedman~\citep{Friedman} function, where a baseline main effect is computed through $b^\ast(X_{i}) = sin(\pi X_{i1}X_{i2}) + 2(X_{i1} - 0.5)^2 + X_{i4} + 0.5X_{i5}$.

\subsubsection{Real-world dataset}
\begin{figure}[h]
\begin{center}
\includegraphics[width=\linewidth]{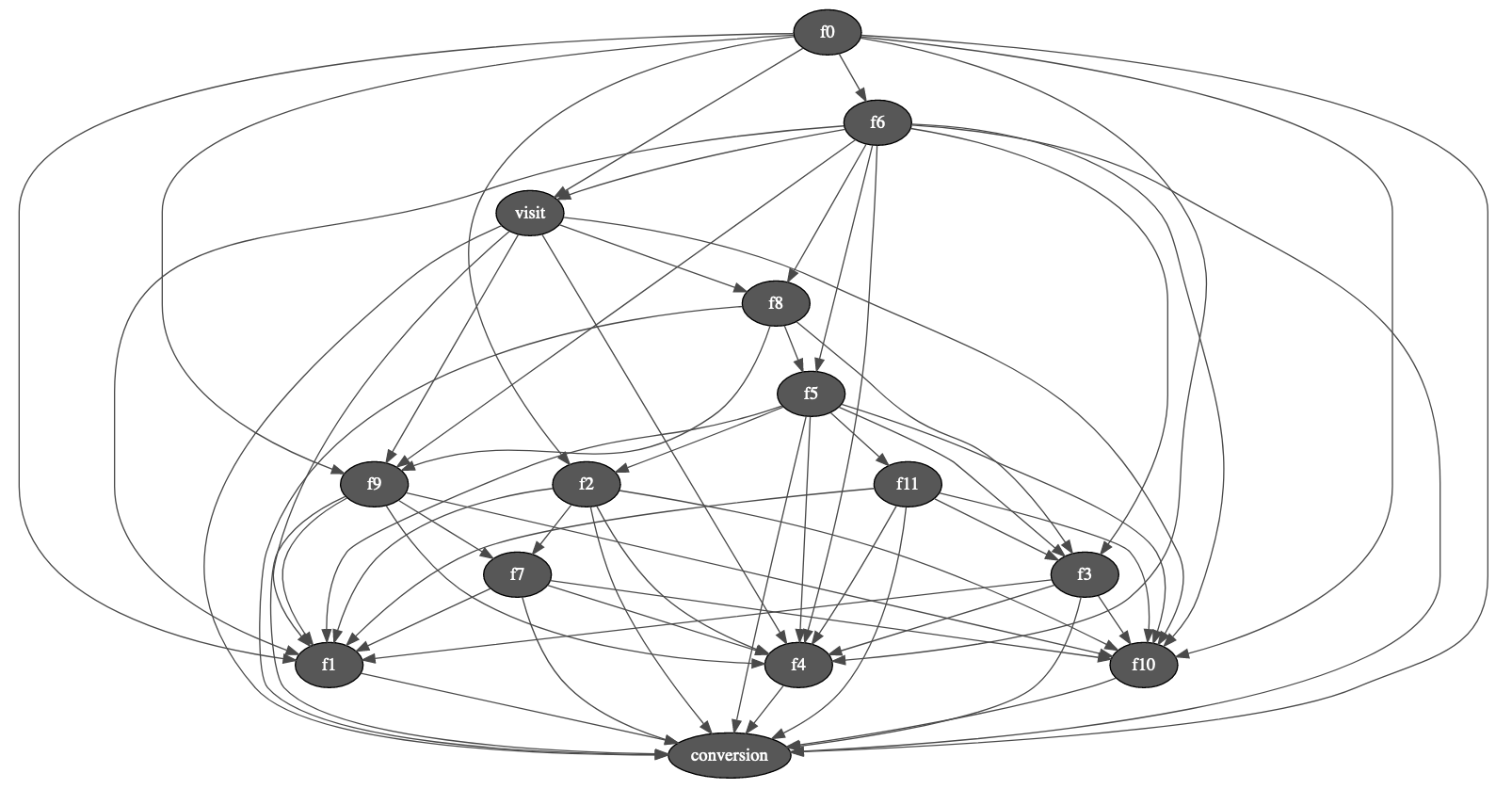}
\end{center}
\caption{Correlation network between confounders, treatment, and outcome from the real-world dataset, CRITEO}
%   \Description{CRITEO Bayesian network}
\label{criteo_bn}
\end{figure}

We use the criteo uplift dataset~\citep{uplift_modeling} as the evaluation of the real-world dataset, which is constructed by collecting data from the incremental test. It randomly divides the people into two categories, whether it is advertised or not. The criteo uplift dataset has 25 million rows, each representing a user with 11 characteristics, a treatment indicator, and two tags (click and conversion). Here we use conversion as the tag we focus on in uplift estimation. Figure~\ref{criteo_bn} shows the results after learning the Bayesian network structure of the dataset.

\subsection{Result}
\subsubsection{synthetic dataset}
In this case, the actual causal effect of features can be calculated easily because the datasets are produced with a certain mechanism The absolute loss (\textit{Abs}) is adopted to measure the deviation between the actual causal effect and the estimated causal effect. As for the prediction accuracy, the mean squared error (\textit{MSE}) is adopted. The proposed method has been compared to traditional models like linear regression (LR), SVR, and XGBoost. 
\begin{figure}[h]
\centering
\includegraphics[width=7.2cm]{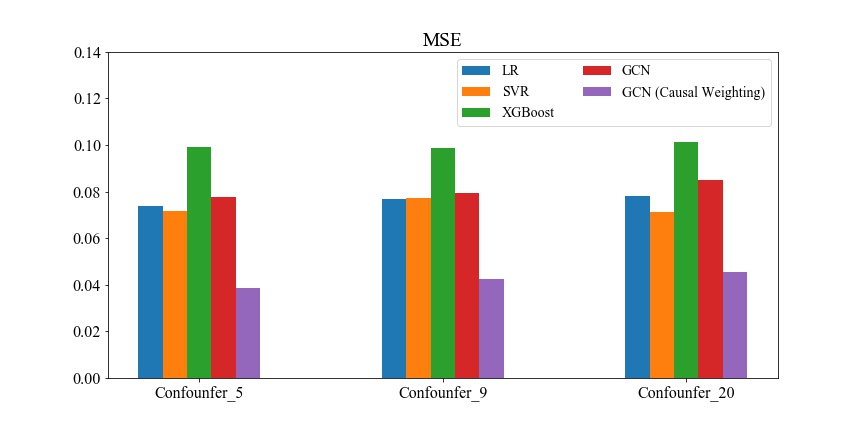}
\caption{Mean squared error for y-prediction accuracy of base methods and ours with the numbers of confounders are 5, 9, and 20, respectively.}
\label{MSE_res}
\end{figure}

\begin{figure}[h]
\centering
\includegraphics[width=7.2cm]{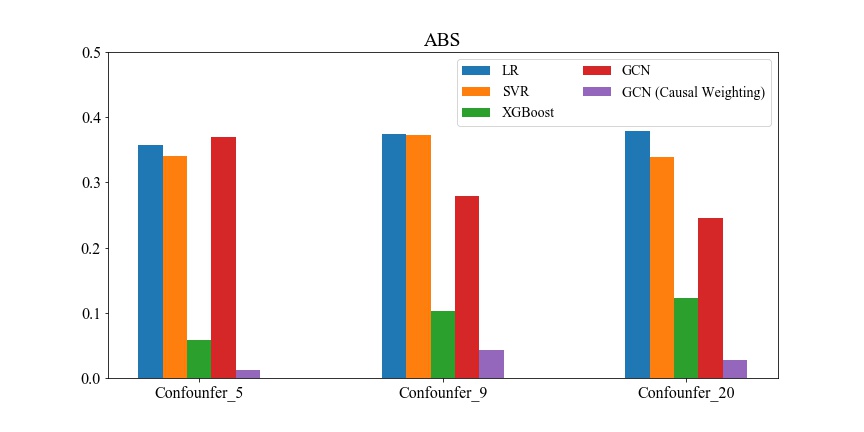}
\caption{Absolute error of ITE of base methods and ours with the numbers of confounders are 5, 9 and 20, respectively.}
\label{ABS_res}
\end{figure}

% \begin{figure}[htbp]
% \centering
% \begin{minipage}[t]{0.46\textwidth}
% \centering
% \includegraphics[width=7.2cm]{MSE.jpg}
% \caption{Mean squared error for y-prediction accuracy of base methods and ours with the numbers of confounders are 5, 9 and 20, respectively.}
% \label{MSE_res}
% \end{minipage}
% \begin{minipage}[t]{0.46\textwidth}
% \centering
% \includegraphics[width=7.2cm]{ABS.jpg}
% \caption{Absolute error of ITE of base methods and ours with the numbers of confounders are 5, 9 and 20, respectively.}
% \label{ABS_res}
% \end{minipage}
% \end{figure}

Figure~\ref{MSE_res} shows that the origin GNN-based model performs similarly to traditional models in the traditional regression task, while the GNN-based model performs much better when combined with the causal weighting. As for uplift modeling estimation, as shown in Figure~\ref{ABS_res}, causal weighting combined architecture has a much more apparent effect. Before combining causal weighting information, GNN based model is slightly better than LR and SVR in the estimation of uplift value but worse than xgboost while achieving a very accurate result when adopting the causal weighting combined architecture. Another result is that GNN based model can have a much more stable performance when the number of confounders increases, which means that it can have a much more robust performance when facing more complex situations.

% \begin{figure}[h]
%   \centering
%   \includegraphics[width=\linewidth]{KL.jpg}
%   \caption{indicates the KL divergence of basic models and ours while the numbers of confounders are 5, 9 and 20, respectively.}
% %   \Description{CRITEO Bayesian network}
% \end{figure}

\subsubsection{Real-world dataset}
In a real-world dataset, the actual causal effect of treatment remains unknown, leading to the abovementioned indicators being inapplicable. As a result, Area Under Uplift Curve (AUUC) is adopted  to measure the performance of an uplifting model on the real-world dataset. AUUC can be calculated as follows:

\begin{equation}
    AUUC(f)=\int_{0}^{1} V(f, x) d x \approx \sum_{k=1}^{n} V(f, k)
\end{equation}

\begin{equation}
    where V(f, k)=\frac{1}{|T|} \sum_{i \in f(\mathcal{D}, k)} {y_{i}^{1}}_{\left[t_{i}=1\right]} -\frac{1}{|C|} \sum_{j \in f(\mathcal{D}, k)} {y_{j}^{1}}_{\left[t_{j}=0\right]}
\end{equation}

Here $f(\mathcal{D}, k)$ can be the k first samples of the dataset when ordered by the prediction of the model $f$, $|T|$ is the number of samples in the treatment group(t=1), and $|C|$ is the number of samples in the control group(t=0).

For certain causal relationships, the higher the AUUC is, the better the uplifting model performs. The AUUC of the baseline models and ours are listed in Table~\ref{auuc_res}.

\begin{table}
\centering
\caption{AUUC for uplifting evaluation and MSE for y-prediction of baseline methods and proposed method}
\label{auuc_res}
\begin{center}
\begin{tabular}{lll}
\multicolumn{1}{c}{\bf Model}  &\multicolumn{1}{c}{\bf AUUC} &\multicolumn{1}{c}{\bf MSE}
\\ \hline \\
LR & 0.4980 & 0.0026\\
SVR &  0.5475 & 0.0037\\
XGBoost &  0.8756 & 0.0025 \\
GCN & 0.5443 & 0.0028 \\
GCN (Causal Weighting) & 0.8807 & 5e-06 \\
\end{tabular}    
\end{center}
\end{table}

Table~\ref{auuc_res} shows that when estimating the uplift value in the real-world dataset, although origin GNN has a similar performance with LR and SVR, it has a better performance than XGBoost when with the causal weighting combined architecture.

\section{Conclusion and Future Work}
In this work, we investigated how to describe causal information in uplift modeling (add conditional average treatment effect (CATE) and build an adjacency matrix using Bayesian network structure learning). In addition, we addressed how to incorporate this causal information into uplift estimations by proposing a framework for uplift modeling that is based on graph neural networks.

Experiments on simulated and real-world datasets reveal that while the origin graph convolutional neural network performs comparably to conventional approaches when directly predicting uplift values, when paired with causal neighbourhood features and causal representation information, it demonstrates exceptional performance in both the prediction job and the uplift estimation task of the target, owing to the GCN's excellent neighbourhood learning features.

It is worthwhile to investigate more methods of characterising causal knowledge in the future. Weighted adjacent  matrices might be seen as a means of guiding graph convolutional neural networks to provide accurate data. Alternatively, it is equally intriguing to investigate the size of the receptive domain of neighbourhood features. A wider receptive domain denotes more information, which might aid us in enhancing the performance of this job in downstream prediction.

\bibliography{reference}
\bibliographystyle{ACM-Reference-Format}

\appendix

\section{Appendix}

\subsection{Code Availability}
The code that supports the findings of this study will be available on GitHub. 

\end{document}